\documentclass[11pt]{article}
\usepackage{times}
\usepackage{latexsym}
\usepackage{amsmath}
\usepackage{subfig, multicol}
\usepackage[all]{xy}
\usepackage[round]{natbib}
\usepackage{url}
\usepackage[T1]{fontenc}
\usepackage{amsfonts}
\usepackage{dsfont}
\usepackage{arydshln}

\begin{document}

\title{A Finnish News Corpus for Named Entity Recognition}
\author{Teemu Ruokolainen$^1$ \and Pekka Kauppinen$^2$ \and Miikka Silfverberg$^2$ \and Krister Lind\'en$^2$}
\date{%
    $^1$University of Helsinki, National Library of Finland, \\ teemu.ruokolainen@iki.fi \\ \vspace{0.5cm}%
    $^2$University of Helsinki,\\ Department of Modern Languages \\ \vspace{0.5cm} %
}

\maketitle

\begin{abstract}
\noindent We present a corpus of Finnish news articles with a manually prepared named entity annotation. The corpus consists of  953 articles (193,742 word tokens) with six named entity classes (organization, location, person, product, event, and date). The articles are extracted from the archives of Digitoday, a Finnish online technology news source. The corpus is available for research purposes. We present baseline experiments on the corpus using a rule-based and two deep learning systems on two, in-domain and out-of-domain, test sets.

\end{abstract}
\sloppy
\section{Introduction}
\label{sec: introduction}

Named entity recognition (NER) is a fundamental textual information extraction task, in which the aim is to locate and classify named entity expressions into pre-defined classes \citep{nadeau2007}. The set of classes typically include such entities as \textit{persons}, \textit{locations}, and \textit{organizations}. While the core NER task as known today was originally proposed already in the late 1990s \citep{grishman1996}, developing novel NER systems and techniques continues to be an active field of research in natural language processing to this day \citep{lample2016,ma2016,gungor2018b,gungor2018a,katiyar2018,sohrab2018,ju2018}.

Development of automatic NER systems requires manually annotated data, that is, a corpus of running text with manually located and classified named entity tags. In addition to being employed to evaluate system performances, the data is necessary for training models with machine learning techniques.\footnote{While there does exist a body of work aiming at learning named entity recognition systems from unannotated data in an unsupervised manner, supervised learning from annotated data has been the prevalent approach in literature.} Such data resources are currently available for multiple European languages in varying domains. For example, consider work on English newswire, biomedical, and historical archive texts \citep{grishman1996,tjong2003,ohta2002,byrne2007}, German newswire and Wikipedia articles \citep{tjong2003,benikova2014}, a wide range of domains such as newswire, prose, and scientific writing in Swedish \citep{gustafson2006}, Hungarian business articles \citep{szarvas2006hungarian}, Spanish and Catalan newswire text \citep{taule2008}, and modern and old French newswire text \citep{rosset2012}. 

In this work, we first present a corpus consisting of Finnish technology related news articles with a manually prepared named entity annotation. The text material was extracted from the archives of Digitoday, a Finnish online technology news source.\footnote{\url{www.digitoday.fi}} The corpus is available for research purposes and can be readily used for development of NER systems for Finnish. We then train/develop three NER systems on the corpus: 
\textsc{FiNER}, a rule-based NER system developed at University of Helsinki, and two recently published neural network architectures \citep{gungor2018b,katiyar2018}. 
In order to evaluate the systems, we create two additional data sets consisting of Digitoday and Wikipedia articles. These sets correspond to in-domain and out-of-domain evaluation sets, respectively.\footnote{The data sets are available through FIN-CLARIN at \url{http://urn.fi/urn:nbn:fi:lb-2019050201}} 
Essentially, \textsc{FiNER}\footnote{The \textsc{FiNER} system and its technical documentation are available at \url{urn.fi/urn:nbn:fi:lb-2018091301}} should be regarded as a companion tool to the published data sets, The reason for its development was two-fold. Firstly, the rule-based system was intended as a consistency check for the manual NE annotations. Secondly, \textsc{FiNER} was intended as a dynamic extension of the gold-standard training data documented in this paper, i.e. a tool for speeding up the annotation of additional training data in new modern Finnish domains.

The rest of the paper is organized as follows. We describe the Digitoday corpus in Section \ref{sec: corpus}. The evaluation sets and conducted experiments are presented in Section \ref{sec: experiments}. Finally, discussion and conclusions on the work are presented in Sections \ref{sec: discussion} and \ref{sec: conclusions}, respectively.

\section{Digitoday Corpus}
\label{sec: corpus}

In this section, we describe the Digitoday corpus in terms of chosen text material, the annotation process, set of named entity classes, and corpus statistics. This corpus is then used as a training/development data for NER systems evaluated on two external evaluation sets described in the experiments in Section \ref{sec: experiments}. 

\subsection{Text}
\label{sec: text}

The text material was extracted from the archives of Digitoday, a Finnish online technology news source. The material covers a variety of technology related topics, such as business, science, and information security.  The material is extracted from articles published in 2014 under the Creative Commons (CC BY-ND-NC 1.0 FI) licence. The extracted text contains 953 articles (193,742 word tokens) in total. 
In addition to raw word forms, the corpus includes meta data describing if the word tokens belong to a headline, an ingress, or an article body. Figures, tables, and respective captions were not included in the corpus. The text was tokenized using the \texttt{hfst-pmatch} toolkit described in \citep{hardwick2015}.

\subsection{Annotation Process}
\label{sec: annotation process}

The annotation was created in two stages as a collaboration between four annotators (A,B,C, and D). Each stage consisted of three parts: annotation, discussion, and refinement. 

In the first stage, annotator A performed a preliminary annotation of the corpus which yielded the first version of the set of named entities and annotation practices described in Section \ref{sec: named entities} and \ref{sec: annotation practices}, respectively.
Subsequently, the set of entities and practices were discussed between annotators A, B, and C, and the annotation was refined correspondingly by Annotator A.  
This first stage was performed during the course of twelve months: the preliminary annotation phase took roughly seven and the refinement phase roughly five months, respectively. During this time annotator A was employed as a part-time employee assigning half of weekly working hours to the annotation project.  In practice, the annotation was performed using a standard spreadsheet software by assigning the tokenized running text into the first column, and the named entities in the second. 

In the second annotation stage, Annotator D joined the project. The task of annotator D was to extend the rule-based \textsc{FiNER} system, the development of which had started at University of Helsinki previously, utilizing the annotated corpus as a development set. As a result of this toolkit development stage spanning over 12 months, annotator D had gathered 1) a set of possible annotation errors given the annotation practices provided by the first author and 2) a set of refinement suggestions to the annotation practices. Subsequent to a discussion between annotators A, B, C , and D, the named entity categories and annotation practices were updated, and the annotation of the complete data was refined for the second and final time by annotator A. 

Descriptions and discussion of named entity classes and annotation practices are presented in Sections \ref{sec: named entities} and \ref{sec: annotation practices}, respectively. A further discussion on how the annotation evolved during the first and second annotation stages is presented in Section \ref{sec: differences between annotation stages 1 and 2}.

\subsection{Named Entities}
\label{sec: named entities}

The set of named entity classes contains the three fundamental entity types, \textit{person}, \textit{organization}, \textit{location}, collectively referred to as the \textit{enamex} since MUC-6 competition \citep{grishman1996}. In addition, our annotation contains the classes \textit{products}, \textit{events}, and \textit{dates}. In what follows, we discuss these classes from two perspectives, namely, their subcategories along with examples and class-specific annotation practices.

\paragraph{Person (PER)}

Marked person names include:

\begin{itemize}

\item[1.] First names: e.g. Sauli, Barack
\item[2.] Family names: e.g. Niinist\"o, Obama
\item[3.] Aliases, nicknames: e.g. DoctorClu, Kim Dotcom
\item[4.] Combinations of 1-3: e.g. Sauli Niinist\"o, Marko ''Fobba'' Forss
\item[5.] Fictional/mythological characters: e.g. Joulupukki (Santa Claus)

\end{itemize}

\paragraph{Location (LOC)}

Marked subcategories of locations include:

\begin{itemize}

\item[1.] Buildings: e.g. Valkoinen talo (the White House), Lasipalatsi, World Trade Center
\item[2.] Cities, towns, city districts: e.g. Helsinki, New York, P\"alk\"ane, Katajanokka
\item[3.] Continents: e.g. Eurooppa (Europe)
\item[4.] Countries, states: e.g. Suomi (Finland), Kalifornia (California)
\item[5.] Geographical areas: e.g. Latinalainen Amerikka (Latin America), Pohjoismaat (Nordic countries), It\"a-Eurooppa (eastern Europe), manner-Kiina (mainland China)
\item[6.] Parks: Huis Ten Boch -teemapuisto, Yosemiten kansallispuisto (Yosemite national park)
\item[7.] Planets, celestial objects: e.g. Mars, Maapallo (Earth), Kuu (Moon)
\item[8.] Seas, lakes, rivers: Atlantti (the Atlantic), Volga

\end{itemize}

\paragraph{Organization (ORG)}

Marked subcategories of organizations include:

\begin{itemize}

\item[1.] Commercial companies: e.g. Nokia, Apple, Time Warner
\item[2.] Commissions: e.g. Kalifornian Public Utilities -komissio 
\item[3.] Communities/groups of people: e.g. Google Orkut, The Kinks, Stop the cyborgs -kansalaisliike
\item[4.] Education, research, and scientific institutes: e.g. Turun Yliopisto, Carnegie Mellon University, Poliisiammattikorkeakoulu (the (Finnish) Police University College), Euroopan avaruusj\"arjest\"o (the European Space Association), Suomen Ilmatieteen laitos (The Finnish Meteorological Institute)
\item[5.] Judicial systems: e.g. Helsingin hovioikeus (Helsinki court of appeals), Yhdysvaltain korkein oikeus (The Supreme Court of the United States), Euroopan Unionin tuomioistuin (The Court of Justice of the European Union)
\item[6.] Law enforcement organizations: e.g. Keskusrikospoliisi (the Finnish National Bureau of Investigation), Yhdysvaltain liittovaltion poliisi (the Federal Bureau of Investigation), Australian poliisi (Australian police force), New Yorkin poliisi (New York police department) 
\item[7.] News agencies/News services/Newspapers/Newsrooms/News sites/News blogs: e.g. Reuters, Helsingin Sanomat, Foss Patents -blogi
\item[8.] Political parties: e.g. Kokoomus (the National Coalition Party)
\item[9.] Public administration: e.g. Suomen hallitus (the Finnish Government), Euroopan Unioni (the European Union), Ulkoministeri\"o (the Finnish Ministry for Foreign Affairs), Tulli (Finnish Customs), Helsingin kaupunki (City of Helsinki)
\item[10.] Sport leagues: e.g. National Hockey League (NHL)
\item[11.] Stock exchange, banks: e.g. New Yorkin p\"orssi (New York Stock Exchange), Suomen Pankki (the Bank of Finland)
\item[12.] Television networks/stations/channels: e.g. MTV3, FOX
\item[13.] Websites (referring to the underlying organization): e.g. Amazon.com, Verkkokauppa.com

\end{itemize}

\paragraph{Product (PROD)}

Marked subcategories of products include:

\begin{itemize}

\item[1.] Artifacts: e.g. IPhone 6, Lumia-puhelin, Hubble-avaruuskaukoputki (Hubble telescope), Compute Stick, Playstation, Snapdragon-suoritin, Watson-supertietokone
\item[2.] Laws/initiatives: e.g. Patriot Act-laki (the Patriot Act law), J\"arke\"a tekij\"anoikeuslakiin -kansalaisaloite (Make sense to copyright law -citizen initiative)
\item[3.] Networks (other than television network): e.g. Tor
\item[4.] Platforms: e.g. Google Play, Kickstarter
\item[5.] Product series/collections: e.g. -sarja, -mallisto
\item[6.] Programming languages: e.g. Java, C++
\item[7.] Projects/Programs/Operations: e.g. Vitja, VideoLan-projekti, Blue Shield -sairausvakuutusohjelma, Operation Pawn Storm
\item[8.] Protocols: e.g. pop3, imap, VPN
\item[9.] Published copies and issues: e.g. Helsingin Sanomat
\item[10.] Services/Platforms: e.g. Apple Store, Google Play, Tor Mail -s\"ahk\"opostipalvelu
\item[11.] Slogans: e.g. ''Don't be evil''
\item[12.] Software: e.g. Windows 10, Trojan-Banker.Win32.Chthonic, Dropbox
\item[13.] Systems: e.g. Alipay-j\"arjestelm\"a
\item[14.] Technologies: e.g. HoloLens-teknologia
\item[15.] Vehicles/Vessels: e.g. Tesla Model S, Opportunity, Kansainv\"alinen avaruusasema (International Space Station), Helsingin Metro
\item[16.] Websites: e.g. Amazon.com-sivusto, Tvkaista.fi-verkko-osoite
\item[17.] Works/Art: e.g. Tuntematon Sotilas, Hurt Locker, Angry Birds, American Idol

\end{itemize}

Version numbers of products are in included in the whole expression if they appear in the immediate context of the product name: for example, consider ''Windows 10'', ''iPhone 6'', ''Android 5.0'', ''Androidin versio 5.0'' (version 5.0 of Android). In contrast, version names are marked even if they appear individually: for example, consider ''Vista'' and ''Lollipop'' referring to ''Windows Vista'' and ''Android Lollipop'', respectively.

\paragraph{Event (EVENT)}

Marked subcategories of events include:

\begin{itemize}

\item[1.] Expos: e.g. CES-messut
\item[2.] Explicitly marked events: e.g. Mobile World-tapahtuma

\end{itemize}

\paragraph{Date (DATE)} 

Marked forms of date expressions include:

\begin{itemize}
\item[1.] 1.10.2016
\item[2.] lokakuussa (in October)
\item[2.] lokakuun aikana (in October)
\item[3.] 1. lokakuuta (1st October)
\item[4.] 1. p\"aiv\"a lokakuuta (1st day of October)
\item[5.] lokakuun 1. (1st October)
\item[6.] lokakuun 1. p\"aiv\"a (1st day of October)
\item[7.] lokakuun 1. p\"aiv\"a (1st day of October)
\item[7.] 2016
\item[8.] vuonna 2016 (in the year 2016)
\item[8.] vuoden 2016 aikana (during the year 2016)
\item[9.] lokakuussa 2016 (in October 2016)
\item[10.] vuoden 2016 lokakuussa (in October 2016)
\item[11.] 1. ja 2. lokakuuta (1st and 2nd October)
\item[12.]  lokakuun 1. ja 2. p\"aiv\"a (1st and 2nd October)
\item[13.] vuosina 2000 ja 2001, vuosien 2000 ja 2001 aikana (during the years 2000 and 2001)
\item[14.] 1.-10. lokakuuta (from 1st to 10th October)
\item[15.] 1. - 10. lokakuuta (from 1st to 10th October)
\item[16.] tammikuusta lokakuuhun (from January to October)
\item[17.] tammi-lokakuussa, tammi-lokakuun aikana (between January and October) 
\item[18.] 2000-2001
\item[19.] vuodesta 2000 vuoteen 2010 (from the year 2000 to 2001)
\item[20.] vuoden 2000 lokakuusta vuoden 2001 tammikuukuuhun (from October 2000 to January 2001)
\item[21.] vuoden 2000 tammikuusta lokakuuhun (from January to October in 2000)

\end{itemize} 

According to our definition employed here, a date is a time expression which can be expressed as a triplet (day, month, year): for example, consider ''3.10.2016'' and ''3. lokakuuta 2016'' (3rd October 2016). To be marked, at least the month or year has to be explicitly specified: for example, consider ''3. lokakuuta'' (3rd October), ''lokakuussa 2016'' (in October 2016), and ''2016''. Therefore, expressions such as ''3. p\"aiv\"a t\"at\"a kuuta'' (3rd of this month) are not markable. Numerical values can be spelled out or expressed using digits.

Expressions such as ''vuonna 2016'' (in the year 2016) and ''3. p\"aiv\"a lokakuuta" (3rd day of October) are marked as a whole. This is because the expressions ''3. p\"aiv\"a lokakuuta'' and ''vuonna 2016'' have the same meaning as ''3. lokakuuta'' and ''2016'', respectively. Months may sometimes have a purely nominal use and are not marked in those instances: for example, consider 'Tammikuu on kylm\"a kuukausi'' (January is a cold month). 

Expressions employing coordination such as ''lokakuun 1. ja 2. p\"aiv\"a'' (1st and 2nd October) and ''vuosina 2000 ja 2001'' (during the years 2000 and 2001) are considered marked as a whole. Similarly, from-to expressions are marked as a single expression: for example, consider ''2000-2010'', ''2000 - 2010'', ''vuodesta 2000 vuoteen 2010'', ''1.-3. lokakuuta'', ''lokakuun ensimm\"aisest\"a p\"aivast\"a kolmanteen''. Adpositions within from-to expressions are marked: for example, consider ''vuosien 2010 ja 2011 aikana'' (during the years 2010 and 2011).

\subsection{Annotation Practices}
\label{sec: annotation practices}

In the following, we address common annotation practices which are shared by all named entity classes.

\paragraph{Expression \textit{-niminen}}
	
Written Finnish regularly employs a type of expression, in which a name is associated with a noun using a suffix ''-niminen'': for example, consider ''Sauli-niminen henkil\"o'' (a person named Sauli) or ''iPhone-niminen puhelin'' (a phone named iPhone). In these cases, we mark the complete expression, that is, 'Sauli-niminen henkil\"o'' and ''iPhone-niminen puhelin'' are marked as single \textit{person} and \textit{product} entities, respectively. We generalize this rule to also cover cases ''-merkkinen'' (of brand) and ''-mallinen'' (of model): for example, consider ''Tesla-merkkinen auto'' (a car of the brand Tesla) and ''Samsung-mallinen puhelin'' (a phone of the model Samsung). This practice was coined in the first annotation stage and was retained in the second.

\paragraph{Derivation and Compounding}

To be marked, each class must appear as a whole and not as a derivation or as a part of a token. For example, ''Suomi'' (Finland) is marked as a location, whereas ''suomalainen'' (Finnish) and ''Suomi-fani'' (a fan of Finland) are not instances of any class.  An important exception to this rule is the set of cases, in which a named entity forms a part of a compound word while the whole compound word refers to that same entity: for example, consider the expression ''Google-ohjelmistoyhti\"o'' (the software company Google) instead of  ''Google'' or ''iPhone-puhelin'' (iPhone phone) instead of ''iPhone''. In these cases, the compound word is marked as a whole. This practice was coined in the first annotation stage and was kept intact in the second.

\paragraph{Coordination}

Written Finnish regularly employs a type of coordination between compound words which share a common part. For example, the expression ''Windows-k\"aytt\"oj\"arjestelm\"a ja Linux-k\"aytt\"oj\"arjestelm\"a'' (Windows operating system and Linux operating system) is written in a more concise manner as ''Windows- ja Linux-k\"aytt\"oj\"arjestelm\"at'' (Windows and Linux operating systems). In the first annotation stage, the coordination cases were treated as separate entities similarly to, for example, \citep{ohta2002}. For example, consider ''\textbf{Windows-} ja \textbf{Linux-k\"aytt\"oj\"arjestelm\"at}''. However, in the second stage we decided to treat them as single entities as this preserves the connection between ''Windows-'' and ''k\"aytt\"oj\"arjestelm\"a'' which would otherwise be lost.
Another related coordination issue addresses such cases as ''Windows 8 ja 10'' (Windows 8 and 10) or ''iPhone 6 sek\"a 6s Plus'' (iPhone 6 as well as 6s Plus). These cases were also marked as single expressions in the second annotation stage. 
Note that the rule generalizes to multiple tokens: for example, consider '\textbf{Windows XP, Vista ja 10}''.

\paragraph{Quotes}

If an entity, or a part of it, is enclosed in quotes, the quotes are included in the entity without exceptions. This practice was coined in the second annotation stage to make the treatment of quotes consistent.

\paragraph{Abbreviations and Acronyms}

Acronyms and abbreviations used as aliases for named entities are marked with their corresponding entity class. If a named entity is immediately followed by its abbrevation or acronym, both are marked as a single entity. For example, consider sentence fragments ''Yhdysvaltain Tiedusteluvirasto (NSA)'' or ''Yhdysvaltain Tiedusteluvirasto NSA''. In these cases, ''Yhdysvaltain Tiedusteluvirasto (NSA)'' and ''Yhdysvaltain Tiedusteluvirasto NSA'' are marked as single organization entities.

\subsection{Differences Between Annotation Stages 1 and 2}
\label{sec: differences between annotation stages 1 and 2}

As described in Section \ref{sec: annotation process}, the annotation was performed in two stages. The first stage was performed by the first annotator and resulted in the first versions of the annotation and the practices described in Section \ref{sec: annotation practices} while the second stage consisted of the refinement of both annotation and practices according to the suggestions of the second annotator. We will next examine how much and how the annotation changed between the two phases. 

As a coarse overview of the differences, Table \ref{tab: statistics stage 1 and 2} shows the counts of each named entity class in the data after both stages. According to this analysis, the annotation changed very little: the class with the largest relative change was organizations with a mere increase of 27 entities corresponding to 0.22 percentage points. This indicates that the annotators in general agreed on the annotation in the vast majority of cases. 

As a more fine-grained examination, we divide the differences between the annotations into 6 types (the number of occurrences in square brackets): 

\paragraph{1. Addition of an entity [32]} The first type consists of cases in which the first annotator had missed an entity. For example, consider the phrase ''Applen toinen perustaja Steve Wozniak'' (''Apple's second founder Steve Wozniak'') where ''Apple'' had not been marked while it clearly refers to the technology company. 

\paragraph{2. Removal of an entity [41]} The second type consists of cases, in which the first annotator had erroneously marked an entity where there should not have been one. For example, according to our annotation practice considering derivation and compounding, ''Android-puhelin'' (''Android phone'') should not be marked as a product since ''Android'' refers to an operating system and not a phone. Nevertheless, ''Android-puhelin'' was incorrectly marked as a product five times in the first annotation stage.

\paragraph{3. An entity's class is changed while its span is left unchanged [11]} The third type corresponds to cases where the first annotator had correctly located an entity but assigned an erroneous class for it. For example, consider the sentence fragment ''Synchronoss on New Jerseyss\"a p\"a\"amajaansa pit\"av\"a yhti\"o'' (''Synchronoss is a company based in New Jersey'') where ''Synchronoss'' had been assigned the class product instead of organization. 

\paragraph{4. An entity's span is changed while its class is left unchanged [53]} The fourth type contains cases where the span of an entity needed to be modified. This is also the most abundant type of change with 53 occurrences. The high number was mainly due to the refinement in the annotation practice which stated that all quotation marks surrounding entities are included in the entity. Consequently, for example, in the sentence fragment 'nimimerkki " Kotijuristi " kertoi' ('alias " Kotijuristi " told'), in the first annotation the quotation marks were not included in the person entity but were subsequently added.

\paragraph{5. A single entity is divided into two or more entities [22]} The fifth type contained cases where the first annotator had erroneously marked two consecutive entities as a single entity. For example, consider ''Microsoftin Windows 7 -k\"aytt\"oj\"arjestelm\"a'' (''Microsoft's Windows 7 operating system'') where the entire fragment was marked as a product instead of a company (''Microsoftin'') followed by a product (''Windows 7 -k\"aytt\"oj\"arjestelm\"a'').

\paragraph{6. Two or more entities are combined into a single entity [15]} The cases in the sixth and final type rose mainly from the changes in the treatment of quotes and coordination as described above in Section \ref{sec: annotation practices}. For example, consider the name 'Andrew ''Weev'' Auernheimer' which according to the refined practice is marked as a single entity but in the first annotation was marked as three names separated by the quotation marks. Similarly, the change in coordination practice which resulted cases such as ''Windows- ja Linux-k\"aytt\"oj\"arjestelm\"at'' and ''Windows 8 ja 10'' to be marked as single entities contributed to this category.

\begin{table}[h!]
\begin{center}
\begin{tabular}{l|ccc} 
\noalign{\smallskip}
class & stage 1 & stage 2 & $\Delta$ \\ 
\hline
\noalign{\smallskip}
ORG & 9099 & 9137 & 38 \\
PER & 2229 & 2214 & -15 \\
LOC & 2039 & 2022 & -17 \\
DATE & 956 & 955 & -1 \\
PRO & 4458 & 4446 & -12 \\
EVENT & 93 & 93 & 0 \\
\hline 
\noalign{\smallskip}
TOTAL & 18874 & 18863 & -11 \\
\end{tabular}
\end{center}
\caption{Counts (second and third column) and their differences (fourth column) in named entity classes after the first and second annotation stages.}
\label{tab: statistics stage 1 and 2} 
\end{table}

\subsection{Nested Annotation}

In general, the marked name expressions can span multiple word tokens -- for example, consider "Sauli Niinist\"o" or ''Nokia Solutions and Networks''. As for the multi-token expressions, it is then possible to employ either a \textit{nested} or \textit{non-nested}  annotation approach. In the nested case, an expression such as ''Helsingin Yliopisto''(''University of Helsinki'') is marked as an organization while its sub-part ''Helsinki'' is marked as a location. Meanwhile, in the non-nested case, only the longer expression (in this example ''Helsingin Yliopisto'') is marked. In what follows, we will refer to the longer entity as the ''top-level entity'' and its sub-part(s) as the ''nested entity''. While perhaps the most well-known named entity corpora employ the non-nested approach \citep{grishman1996,tjong2003}, this results in a loss of information motivating the use of the nested approach \citep{ohta2002,byrne2007,benikova2014}.

The annotation described so far (resulting after the first and second stages described in Section \ref{sec: annotation process}) consisted of only the top-level entities, that is, only the longest expressions were marked. We then extended the annotation to include also nested entities. The added annotation effort required was rather small because, by definition, the potential nested entities necessarily have to appear inside the top-level entities with longer spans.\footnote{It should be noted that entities can, in general, overlap in two ways, namely, by being nested or by crossing. In the latter, two entities overlap but neither is contained in another. However, such cases were not encountered in the data.} In consequence, in order to produce the nested annotation, one is required only to consider the already marked entities, that is, a fraction of all word tokens in the corpus. Extending the annotation in such a manner required only roughly a few days of additional work. In practice, we assigned the nested entities to a third column of the standard spreadsheet software next to the word tokens and top-level entities in the first and second, respectively. 

The nested annotation followed four basic practices. First, a nested entity can not be of same length as its top-level counterpart. This condition simply follows from the fact that an entity can not have two valid interpretations simultaneously. For example, given context, an instance of ''Nokia'' can not be both a company and a city nor ''Ashley Madison'' both a company and a person at the same time. Second, we did not allow a nested entity to be of the same class as its top-level counterpart. For example, ''Microsoft Research'' would be marked as a single non-nested organization entity with two tokens (in contrast to ''Microsoft'' forming a single-token child organization entity within it). This was because the majority of such cases in the data appeared to contain rather redundant information exemplified by ''York'' nested within ''New York'' or ''Nokia'' nested within ''Nokia Solutions and Networks''. Third, a top-level entity was allowed to have more than one nested entity of one or more classes (but not of the same class as the parent entity). Fourth, we allowed more than two levels of annotation, that is, a nested entity having a nested entity. However, no such instances were encountered.  

Table \ref{tab: counts of classes} shows the number of class occurrences on the top-level and nested annotation. The total number of marked entities increased by 4.3\% from 18,863 to 19,667. The largest increase after adding the nested annotation took place in locations, the number of which increased by 410 from 2022 to 2432. Nevertheless, we note that the number of nested entities is quite small compared with the number of all entities. Of the 19667 entities, only 804 (4.1\%) were embedded within another entity. For comparison, in the German NoStad-D \citep{benikova2014} corpus consisting of online news and Wikipedia text and the biomedical GENIA corpus \citep{ohta2002}, 7.7\% and 17\% of all entities were embedded within another entity.
\footnote{Note that the GENIA annotation allowed the top-level and nested entities to be of the same class increasing the nested/all-ratio. We are not able to say if this is the case with NoSta-D based on \citep{benikova2014}.} Finally, Table \ref{tab: nested statistics} presents how the nested entities were distributed within the top-level classes. As can be seen, only top-level entities belonging to organizations, products, and events included a nested entity at least once.

\begin{table}[h!]
\begin{center}
\begin{tabular}{l|cccccc} 
class & top-level & \% & nested &  \%  & all & \% \\ 
\hline
\noalign{\smallskip}
ORG & 9137 & 48.43 & 279 & 34.70 & 9416 & 47.87 \\
PER & 2214 & 11.74 & 113 & 14.05 & 2327 & 11.83 \\
LOC & 2022 & 10.72 & 410 & 51.00 & 2432 & 12.37 \\
PRO & 4446 & 23.55 & 2 & 0.25 & 4448 & 22.60 \\
DATE & 955 & 5.06 & 0 & 0.00 & 955 & 4.86 \\
EVENT & 93 & 0.49 & 0 & 0.00 & 93 & 0.47 \\
\hline
\noalign{\smallskip}
TOTAL & 18867 & 100.00 & 804 & 100.00 & 19671 & 100.00 \\
\end{tabular}
\end{center}
\caption{Counts of named entity classes in the corpus.}
\label{tab: counts of classes} 
\end{table}

\begin{table}[h!]
\begin{center}
\begin{tabular}{l|cccccc|c} 
 & ORG & PRO & PER & LOC & DATE & EVENT & NONE \\ 
\hline
\noalign{\smallskip}
ORG & 0 & 0 & 90 & 398 & 0 & 0 & 8649 \\
PRO & 276 & 0 & 21 & 6 & 0 & 0 & 4143 \\
PER & 0 & 0 & 0 & 0 & 0 & 0 & 2214 \\
LOC & 0 & 0 & 0 & 0 & 0 & 0 & 2022 \\
DATE & 0 & 0 & 0 & 0 & 0 & 0 & 955 \\
EVENT & 3 & 0 & 2 & 6 & 0 & 0 & 82 \\
\hline
\noalign{\smallskip}
TOTAL & 276 &  0 & 113 & 404 & 0 & 0 &  18065 \\
\end{tabular}
\end{center}
\caption{Counts of nested entities. The rows and columns correspond to the top-level and nested entities, respectively. For example, a person entity was found 90 times within an organization entity. The column entity 'NONE' refers to the number of top-level entities with no nested entity. For example, there were 8649 occurrences of top-level organization entities without a nested entity.}
\label{tab: nested statistics} 
\end{table}

\subsection{File Format}

The data is presented in a three-column file format using the standard BIO notation with an empty line separating sentences. This format was employed in the CoNLL-2003 shared task (apart from using three columns instead of two to accommodate the nested entities). For example, consider the following sentence ''New York Times vertasi HoloLensi\"a ja Google Glassia.'' (''New York Times compared HoloLens and Google Glass.'') 

\begin{center}
\begin{tabular}{lcc} 
New &  B-ORG &  B-LOC \\
York & I-ORG &  I-LOC \\
Times & I-ORG & O \\
vertasi & O & O \\
HoloLensi\"a & B-PRO & O \\
ja & O & O \\
Google & B-PRO & B-ORG \\
Glassia & I-PRO & O \\
. & O & O \\
\end{tabular}
\end{center}

\subsection{Inter-Annotator Agreement}
\label{sec: iaa}

In order to test the consistency and quality of the annotation guideline presented in Sections \ref{sec: named entities} and \ref{sec: annotation practices}, we measured inter-annotator agreement. We enlisted an additional annotator who, based on the annotation guideline, independently annotated 762 uniformly sampled sentences (10,007 tokens in total) from the Digitoday test set which is described in Section \ref{sec: digitoday test set}. This gave us a  second set of annotations for this data set which we then compared with the pre-existing annotations
described in Section \ref{sec: annotation process}. 
 The pre-existing annotations in this portion of the Digitoday test set contain 910 named entities in total and contain examples from each entity type which is present in the corpus.

\paragraph{Measuring Agreement} To compare annotations, we used Cohen's $\kappa$ \citep{Cohen (1960)} which is a measure for inter-annotator agreement commonly applied in natural language processing \citep{Arstein (2008)}. The $\kappa$ metric in Equation~\ref{eq: kappa} measures agreement between two independent annotators I and II on a classification task.  Here $p_o$ is the so called \textit{observed agreement} which is computed as the relative frequency of elements which were annotated identically by annotators I and II (this is essentially accuracy). The quantity $p_e$ is the probability of \textit{chance agreement} between annotators I and II. It is given by Equation \ref{eq: chance-agreement}, where $N$ denotes the number of elements in the data set and the quantities $n_{k_I}$ and $n_{k_{II}}$ denote the counts of elements classified into the class $k$ by I and II, respectively.

\begin{equation}
    \kappa \equiv 1 - \frac{1 - p_o}{1 - p_e}\label{eq: kappa}
\end{equation}

\begin{equation}
    p_e = \frac{\sum_k n_{k_I} \cdot n_{k_{II}}}{N^2}\label{eq: chance-agreement}
\end{equation}

We apply Cohen's $\kappa$ for named entity recognition by recasting the task as a classification task in the following way. First we form the complete set of named entities annotated in the data set.\footnote{These are instances of named entities, not types. For example,  \textit{New York} may occur several times in the corpus. Each occurrence is considered a separate named entity.} These are then labelled with the entity types assigned by each annotator (for example \texttt{LOC} or \texttt{ORG}). There may exist entities which were not annotated by both I and II. Therefore, we additionally introduce a label \texttt{O} which is used when an expression was not identified as a named entity by one of the annotators. Finally, we compute the $\kappa$ metric for the entity labels. 

Only complete matches are considered when computing $p_o$. For example, if annotator I annotated the expression \textit{The New York Post} as \texttt{ORG} and II instead annotated its sub-sequence \textit{New York Post}, then these are treated as two distinct entities $e_1$ and $e_2$. In this case, the class labels for $e_1$ and $e_2$ assigned by annotator I will be \texttt{ORG} and \texttt{O} and the labels assigned by annotator II will be \texttt{O} and \texttt{ORG}, respectively. Since partial matches between annotators are considered to be errors, this metric is quite strict and is unlikely to exaggerate the agreement between annotators \citep{Arstein (2008)}.

\paragraph{Results} For top-level entities, the measured inter-annotator agreement is $0.79$. Although it is difficult to interpret the $\kappa$ metric in general, an inter-annotator agreement score of $0.79$ can be considered strong \citep{Landis1997}. For all entities, including nested entities, the agreement score is slightly lower at $0.76$. However, even this agreement score can be considered to be strong.

\paragraph{Analysis} Approximately 59\% of all discrepancies between top-level annotations involved the product category \texttt{PRO}. Cases where one of the annotators omitted a product entity and the other one labelled it account for $41\%$ of all discrepancies. Most of these were simple omissions by the external annotator. Additionally, confusion between the product and organization categories was prevalent (15\% of all observed differences). Detailed analysis showed that the confusions mostly stems from websites, application names and company names. Frequently, a name like \textit{YouTube} or \textit{Spotify} can refer both to a company, which should be labelled \texttt{ORG}, and a website or application, which should be labelled \texttt{PRO}. It can be difficult to determine the correct label without access to complete articles (as explained above, the sentences in the experiment were sampled from the corpus). An additional large group of differences concerns identification of the span of organization names. For example, the external annotator labelled \textit{Uutistoimisto Reuters} (the Reuters news agency) as an organization, whereas, the pre-existing annotation only includes the company name \textit{Reuters} in the entity and excludes the qualifying expression \textit{Uutistoimisto} (news agency). However, this problem occurred only when the qualifying expression was placed at the beginning of a sentence where it was capitalized. This lead to it mistakenly being identified as part of the proper name.

For nested entities, most of the differences stemmed from annotators failing to identify a location or organization which is part of a product or organization name. For example, the external annotator correctly identified \textit{Gainesville} in \textit{The Gainesville Sun} as a location, whereas it was not present in the pre-existing corpus annotation. Consequently, most of these discrepancies reflect errors in the annotation process rather than the quality of the annotation guideline. When entities were missing from the corpus annotation, they were added. 

\section{Experiments}
\label{sec: experiments}

In this section, we present experiments conducted employing the corpus described so far. We utilize the corpus as training data to develop one rule-based and two statistical NER systems, and create two new data sets to use as evaluation data. We describe the data, the NER systems, and the evaluation measures in Sections \ref{sec: data}, \ref{sec: ner systems}, and \ref{sec: evaluation}, respectively. Finally, the obtained results are presented and discussed in Section \ref{sec: results}.

\subsection{Data}
\label{sec: data}

In what follows, we discuss the data used for training and development of the NER systems in Section \ref{sec: training data} and the in-domain and out-domain test sets in Sections \ref{sec: digitoday test set} and \ref{sec: wikipedia test set}, respectively.

\subsubsection{Training Data}
\label{sec: training data}

As presented in Section \ref{sec: data}, the annotated corpus consists of 193,742 word tokens in total. We form training and development sets of 183,552 and 10,190 word tokens, respectively.

\subsubsection{Digitoday Test Set}
\label{sec: digitoday test set}

The first test data set consists of articles extracted from Digitoday archives published in 2015, as opposed to the training data consisting of Digitoday articles published in 2014. This set corresponds to an in-domain test set. After preprocessing the text as described in Section \ref{sec: text}, the data contains 240 articles (46,363 word tokens). The data set was annotated by replicating the annotation process described in Section \ref{sec: corpus}: the text was first annotated by a single annotator following the annotation practices described in Section \ref{sec: annotation practices}, after which correction suggestions were provided by a second annotator, and annotation was formed subsequent to a discussion session between the annotators. As presented in Table \ref{tab: statistics stage 1 and 2 digitoday test set}, the annotation changed very little between stages 1 and 2 and, therefore, we do not provide a more in-depth analysis. The final counts of classes after adding the nested entities are presented in Table \ref{tab: digitoday test class counts}.

\begin{table}[h!]
\begin{center}
\begin{tabular}{l|ccc} 
\noalign{\smallskip}
class & stage 1 & stage 2 & $\Delta$ \\ 
\hline
\noalign{\smallskip}
ORG & 1878 & 1879 & 0 \\
LOC & 505 & 511 & 6 \\
PER & 406 & 406 & 0 \\
DATE & 237 & 238 & 1 \\
PRO & 1084 & 1072 & -12 \\
EVENT & 17 & 18 & 1 \\\hline 
\noalign{\smallskip}
TOTAL & 4127 & 4123 & -4 \\
\end{tabular}
\end{center}
\caption{Counts of named entity classes in the Digitoday test data after the first and second annotation stages.}
\label{tab: statistics stage 1 and 2 digitoday test set} 
\end{table}

\begin{table}[h!]
\begin{center}
\begin{tabular}{l|cccccc} 
class & top-level & \% & nested &  \%  & all & \% \\ 
\hline
\noalign{\smallskip}
ORG & 1879 & 45.55 & 154 & 56.41 & 2032 & 46.22 \\
LOC & 511 & 12.39 & 90 & 32.97 & 601 & 13.67 \\
PER & 406 & 9.85 & 27 & 9.89 & 433 & 9.85 \\
DATE & 238 & 5.77 & 2 & 0.73 & 240 & 5.46 \\
PRO & 1072 & 26.00 & 0 & 0.00 & 1072 & 24.39 \\
EVENT & 18 & 0.44 & 0 & 0.00 & 18 & 0.41 \\
\hline
\noalign{\smallskip}
TOTAL & 4123 & 100.00 & 273 & 100.00 & 4396 & 100.00 \\
\end{tabular}
\end{center}
\caption{Counts of top-level and nested entities in the Digitoday test data.}
\label{tab: digitoday test class counts}.
\end{table}

\subsubsection{Wikipedia Test Set}
\label{sec: wikipedia test set}

The second test data set consists of articles extracted from the Finnish Wikipedia archives.\footnote{\url{https://dumps.wikimedia.org/fiwiki/latest/fiwiki-latest-pages-articles.xml.bz2} downloaded 1.2.2018.} This set corresponds to an out-of-domain test set. After preprocessing, the data contains 83 articles (49,752 word tokens). As with the Digitoday test set, the Wikipedia articles were annotated by replicating the annotation process described in Section \ref{sec: corpus}. The counts of entity classes are presented in Table \ref{tab: statistics stage 1 and 2 wikipedia test set}. Again, the annotation changed very little between the stages and, therefore, we do not provide a more in-depth analysis. The final counts of classes after adding the nested entities are presented in Table \ref{tab: wikipedia test class counts}.

As expected, the articles consider a wide range of topics including (but not limited to) sports, music, nature, and technology. While we are not able to provide a detailed review of the Finnish Wikipedia contents here, we note that there are some prominent biases towards certain topics and article types. In particular, 26 out of the 83 randomly selected articles (31\%) are biographies of people known either globally or locally in Finland. Therefore, it is not surprising that, in the Wikipedia data, person names are the most frequent entity class. This is in contrast to Digitoday articles which are dominated by mentions of organizations and products. In addition, the topical distribution of the Wikipedia data appears to be skewed towards a few particular topics with relatively high frequencies such as music (10 out of 83 articles) and sports (9 our of 83 articles). Finally, the Wikipedia articles appear to have a higher named entity density  (roughly 127 name mentions per 1000 word tokens) compared with Digitoday (95 per 1000 word tokens).

\begin{table}[h!]
\begin{center}
\begin{tabular}{l|ccc} 
\noalign{\smallskip}
class & stage 1 & stage 2 & $\Delta$ \\ 
\hline
\noalign{\smallskip}
PER & 1666 & 1666 & 0 \\
DATE & 997 & 1007 & 10 \\
LOC & 1277 & 1292 & 15 \\
ORG & 1031 & 1053 & 22 \\
PRO & 656 & 657 & 1 \\
EVENT & 176 & 156 & -20 \\
\hline 
\noalign{\smallskip}
TOTAL & 5803 & 5831 & 28 \\
\end{tabular}
\end{center}
\caption{Counts of named entity classes in the Wikipedia test data after the first and second annotation stages.}
\label{tab: statistics stage 1 and 2 wikipedia test set} 
\end{table}

\begin{table}[h!]
\begin{center}
\begin{tabular}{l|cccccc} 
class & top-level & \% & nested &  \%  & all & \% \\ 
\hline
\noalign{\smallskip}
PER & 1666 & 28.57 & 131 & 25.79 & 1797 & 28.35 \\
DATE & 1007 & 17.27 & 22 & 4.33 & 1029 & 16.23 \\
LOC & 1292 & 22.16 & 312 & 61.42 & 1604 & 25.30 \\
ORG & 1053 & 18.06 & 40 & 7.87 & 1093 & 17.24 \\
PRO & 657 & 11.27 & 3 & 0.59 & 660 & 10.41 \\
EVENT & 156 & 2.68 & 0 & 0.00 & 156 & 2.46 \\\hline
\noalign{\smallskip}
TOTAL & 5831 & 100.00 & 508 & 100.00 & 6339 & 100.00 \\
\end{tabular}
\end{center}
\caption{Counts of top-level and nested entities in the Wikipedia test data.}
\label{tab: wikipedia test class counts}.
\end{table}

\subsection{Evaluation}
\label{sec: evaluation}

We follow the classic CoNLL-2003 shared task \citep {tjong2003} and evaluate the systems using \textbf{F1-score} which is the harmonic mean of \textbf{precision} (the number of correctly recognized entities divided by the number of all recognized entities) and \textbf{recall} (the number of correctly recognized entities divided by the number of all annotated entities in data). We compute the precision, recall, and F1 measures for each entity class (PER, LOC, ORG, PRO, EVENT, DATE) individually and over all classes to assess the overall performance of each system. Moreover, the measures over all classes can, in general, be obtained in two different ways using either \textbf{macro} or \textbf{micro} averaging. In macro averaging, we first compute precision, recall, and F1 score for each class individually and return their averages. In micro averaging, the precision, recall, and F1 score are computed after first adding up all recognition results. If each class is roughly of equal size, then micro and macro averages are roughly the same. The CoNLL-2003 evaluation employs the micro averaging approach. The evaluations are performed on all entities as well as on top-level entities only.

\subsection{Named Entity Recognizers}
\label{sec: ner systems}

In this section, we describe the three named entity recognizers employed in the experiments, namely, the rule-based \textsc{FiNER} system and two neural network architectures \citep{gungor2018b,sohrab2018}.

\subsubsection{\textsc{FiNER}}
\label{sec: finer}

The Finnish Named Entity Recognizer (\textsc{FiNER}) is a rule-based named entity recognition tool for Finnish. The toolkit is freely available for research and commercial uses. Its pipeline utilizes a combination of morphological analysers/taggers and a set of pattern-matching rules for identifying and classifying proper names and other expressions in plain text input. As its pattern-matching engine, \textsc{FiNER} utilizes the \texttt{hfst-pmatch} toolkit \citep{hardwick2015}. 

The first versions of \textsc{FiNER}, including the basic architecture of the NER rule set, were developed during 2011--2013 at University of Helsinki. The original name hierarchy was based on that employed by the Swedish Named Entity Recognizer \citep{kokkinakis2003} comprising three categories for names (locations, organizations, and people) and two additional categories for temporal and numerical expressions. 
The work on \textsc{FiNER} resumed in late 2016, following the creation of the manually annotated data set described in Section \ref{sec: annotation process}. The data set allowed for a more systematic and rigorous development of the system as well as the HFST toolkit and its \texttt{hfst-pmatch} implementation. Major modifications and improvements that followed included the introduction of two new main categories (products and events), addition of new subcategories to existing categories, new methods of matching foreign names, nested annotations, more extensive use of context restraints for delimiting names, and the introduction of a capture mechanic that allowed the reuse of matched input segments in other patterns.

The reason to develop \textsc{FiNER} was two-fold. Firstly, the rule-based system was intended as a consistency check for the manual NE annotations, i.e. when we annotated the training data, we discussed the nature and shape of the entities and encoded them in a set of rules. The rules were applied to the training data and inconsistent annotations were spotted and further discussed. Secondly, \textsc{FiNER} was intended as a dynamic extension of the gold-standard training data documented in this paper. Ideally, \textsc{FiNER} was intended to be used for annotating additional data in new modern Finnish domains speeding up future efforts to create more gold-standard training data. To ensure this, \textsc{FiNER} has been tested and used on various domains such as general news material, museum object meta-data descriptions, court rulings, various Wikipedia articles (except the ones tested in the test data), etc. For this purpose, lists of well-known named entities in each category have also been collected and made into gazetteers. Many elements of these gazetteers have made it into the morphological analyser \textsc{OMorFi}\footnote{https://www.kielipankki.fi/tools/omorfi/} and are there for any system capable of taking such additional linguistic analysis and semantic categories into account. 

One of the strong features of \textsc{FiNER} through \texttt{hfst-pmatch} is its capability to dynamically adapt to new domains through the capture mechanism. New NEs in defining contexts are automatically added to the gazetteers of FiNER when FiNER recognizes them. The new NEs are then used for annotating their remaining instances in the text being processed. 

\subsubsection{Neural Network Architectures}

The state-of-the-art in statistical NER is currently yielded by recurrent neural network architectures \citep{lample2016,ma2016}. These approaches encode orthographic (what the word looks like) and distributional (in which contexts the word occurs in corpus) evidence using character-based word representation models \citep{ling2015,chiu2016} and distributional representations \citep{mikolov2013,pennington2014}, respectively. A combination of these representations is then fed to a bi-directional long short term-memory (biLSTM) network to model word contexts. Finally, on top of the biLSTM, the architectures employ a conditional random field (CRF) model to capture the strong dependencies between adjacent IOB-encoded labels (e.g. I-LOC can not be preceded by O). 

As shown by \citet{lample2016} and \citet{ma2016}, these approaches yield state-of-the-art accuracy without the need for hand-crafted features or gazetteers employed in previous state-of-the-art statistical models \citep{finkel2005}. In addition, it was recently shown by \citet{gungor2018a,gungor2018b} that, especially for morphologically rich languages such as Turkish and Finnish, using additional word representations based on linguistic properties of the words (encoded as the output of a morphological analyser) improves the performance further compared with using only representations based on the surface forms of words. Despite their success, however, these architectures have an evident shortcoming in that they only consider the top-level NE annotation while ignoring the nested entities during both training and prediction. Therefore, a recent body of work has extended the approach to handle also the nested entities using, for example, exhaustive search \citep{sohrab2018}, hypergraph representations \citep{katiyar2018}, layers of biLSTMs \citep{ju2018}, and transitions \citep{wang2018}. 

In our experiments, we present results using the architectures of \citet{gungor2018b} and \citet{sohrab2018}. The model of \citet{gungor2018b} (referred from now on as \textsc{G\"ung\"or-NN}) is trained using top-level entities only and no nested entities are predicted on the test sets, whereas the model of \citet{sohrab2018} (referred to as \textsc{Sohrab}) is trained on all entities and predicts nested entities for the test sets. The distributional word representations of both models are initialized using the pretrained word embeddings provided by the Finnish Internet Parsebank project \citep{kanerva2014} learned from a corpus of 1.5 billion word tokens in 116 million sentences using the word2vec software \citep{mikolov2013}.\footnote{The pretrained word embeddings can be obtained directly from \url{http://bionlp-www.utu.fi/fin-vector-space-models/fin-word2vec.bin}} The morphological analyses used by \textsc{G\"ung\"or-NN} are obtained using the FinnPos system \citep{silfverberg2016}.\footnote{We employ the pretrained FinnTreeBank tagger available at \url{https://github.com/mpsilfve/FinnPos}}  The use of morphological analysis improves the F1-score of the model on development set by 0.75 (89.10 versus 88.35). The original implementation of \textsc{Sohrab-NN} was extended to also leverage the output of a morphological analyser using the same approach as \textsc{G\"ung\"or-NN}. This improved the F1-score of the model on development set by 1.15 (86.06 versus 84.91).

\subsection{Results}
\label{sec: results}

The results obtained for the Digitoday test set are presented in Tables \ref{tab: results top-level+nested digitoday} (all entities) and \ref{tab: results top-level digitoday} (top-level entities only). The highest overall performance was yielded by the rule-based  \textsc{FiNER} system with F1-scores of 85.20 (all entities) and 86.82 (top-level entities only). The improvement in performance over \textsc{G\"ung\"or-NN}and \textsc{Sohrab-NN} is statistically significant (with confidence level 0.95) based on the standard 2-sided Wilcoxon signed-rank test performed on 20 randomly divided, non-overlapping subsets of the complete test set). The highest overall recall was yielded by \citep{gungor2018b} with recall values of 80.73 (all entities) and 85.62 (top-level entities only).

\begin{table}[t!]
\begin{small}
\begin{center}
\setlength\tabcolsep{5pt}
\begin{tabular}{l|ccc|ccc|ccc}
&  \multicolumn{3}{c}{\textsc{FiNER}} &  \multicolumn{3}{c}{ \textsc{G\"ung\"or-NN}}  &  \multicolumn{3}{c}{ \textsc{Sohrab-NN}}  \\ 
class & pre & rec & F1 & pre & rec & F1 & pre & rec & F1 \\
\hline
\noalign{\smallskip}
DATE & 97.51 & 97.92 & 97.71 &	94.47 & 92.50 & 93.47 &	100.00 & 88.75 & 94.04 \\
EVENT & 100.00 & 100.00 & 100.00 &	100.00 & 66.67 & 80.00 &	100.00 & 50.00 & 66.67 \\
LOC & 93.55 & 89.35 & 91.40 &	91.39 & 81.20 & 85.99 &	89.67 & 85.19 & 87.37 \\
ORG & 93.81 & 80.51 & 86.65 &	86.90 & 83.56 & 85.20 &	89.16 & 80.95 & 84.86 \\
PER & 88.28 & 78.29 & 82.99 &	78.03 & 78.75 & 78.39 &	74.14 & 74.83 & 74.48 \\
PRO & 82.49 & 71.18 & 76.41 &	74.98 & 73.51 & 74.23 &	0.56 & 64.55 & 71.67 \\
\hline
\noalign{\smallskip}
TOTAL & \bf 90.79 & 80.25 & \bf 85.20 &	84.04 & \bf 80.73 & 82.35 &	86.30 & 77.23 & 81.51 \\
\end{tabular}
\end{center}
\end{small}
\caption{Precision, recall, and F1-scores for all named entities in the Digitoday test set. The highest overall precision, recall, and F1 values are bolded.}
\label{tab: results top-level+nested digitoday}
\end{table}

\begin{table}[t!]
\begin{small}
\begin{center}
\setlength\tabcolsep{5pt}
\begin{tabular}{l|ccc|ccc|ccc}
&  \multicolumn{3}{c}{\textsc{FiNER}} &  \multicolumn{3}{c}{ \textsc{G\"ung\"or-NN}}  &  \multicolumn{3}{c}{ \textsc{Sohrab-NN}}  \\ 
class & pre & rec & F1 & pre & rec & F1 & pre & rec & F1 \\
\hline
\noalign{\smallskip} 
DATE & 97.92 & 98.74 & 98.33 &	94.47 & 93.28 & 93.87 &	100.00 & 89.50 & 94.46 \\
EVENT & 100.00 & 100.00 & 100.00 & 	100.00 & 66.67 & 80.00 &	100.00 & 50.00 & 66.67 \\
LOC & 92.53 & 94.52 & 93.51 &	87.83 & 91.78 & 89.76 &	85.03 & 86.69 & 85.85 \\
ORG & 93.48 & 85.57 & 89.35 &	86.90 & 90.42 & 88.62 &	87.18 & 82.59 & 84.82 \\
PER & 87.76 & 83.00 & 85.32 &	78.03 & 83.99 & 80.90 &	73.33 & 78.57 & 75.86 \\
PRO & 82.49 & 71.18 & 76.41 &	74.98 & 73.51 & 74.23 &	80.65 & 64.55 & 71.71 \\
\hline
\noalign{\smallskip}
TOTAL & \bf 90.41 & 83.51 & \bf 86.82 &	83.59 & \bf 85.62 & 84.59 &	84.59 & 78.27 & 81.31 \\
\end{tabular}
\end{center}
\end{small}
\caption{Precision, recall, and F1-scores for top-level named entities in the Digitoday test set. The highest overall precision, recall, and F1 values are bolded.}
\label{tab: results top-level digitoday}
\end{table}

The results obtained for the Wikipedia test set are presented in Tables \ref{tab: results top-level+nested wikipedia} (all entities) and \ref{tab: results top-level wikipedia} (top-level entities only). The highest overall performance was again yielded by \textsc{FiNER} with F1-scores of 79.91 (all entities) and 78.31 (top-level entities only). The improvement over \textsc{G\"ung\"or-NN} and \textsc{Sohrab-NN} is statistically significant. In addition to F1-score, the \textsc{FiNER} system also yielded the highest precision and recall values over all classes.

\begin{table}[t!]
\begin{small}
\begin{center}
\setlength\tabcolsep{5pt}
\begin{tabular}{l|ccc|ccc|ccc}
&  \multicolumn{3}{c}{\textsc{FiNER}} &  \multicolumn{3}{c}{ \textsc{G\"ung\"or-NN}}  &  \multicolumn{3}{c}{ \textsc{Sohrab-NN}}  \\ 
class & pre & rec & F1 & pre & rec & F1 & pre & rec & F1 \\
\hline
\noalign{\smallskip}
DATE & 98.53 & 97.47 & 98.00 &	80.95 & 51.21 & 62.74 &	99.52 & 60.45 & 75.21 \\
EVENT & 70.87 & 57.69 & 63.60 &	13.64 & 1.92 & 3.37 &	0.00 & 0.00 & 0.00 \\
LOC & 88.42 & 77.62 & 82.67 &	75.86 & 72.88 & 74.34 &	77.99 & 57.67 & 66.31 \\
ORG & 79.86 & 50.41 & 61.81 &	47.43 & 29.55 & 36.41 &	36.05 & 26.72 & 30.69 \\
PER & 92.03 & 77.80 & 84.32 &	86.09 & 66.83 & 75.25 &	72.30 & 43.57 & 54.37 \\
PRO & 73.80 & 49.09 & 58.96 &	30.28 & 40.61 & 34.69 &	28.36 & 31.36 & 29.78 \\
\hline
\noalign{\smallskip}
TOTAL & \bf 88.66 & \bf 72.74 & \bf 79.91 &	67.46 & 55.07 & 60.64 &	63.79 & 44.63 & 52.52 \\
\end{tabular}
\end{center}
\end{small}
\caption{Precision, recall, and F1-scores for all named entities in the Wikipedia test set. The highest overall precision, recall, and F1 values are bolded.}
\label{tab: results top-level+nested wikipedia}
\end{table}

\begin{table}[t!]
\begin{small}
\begin{center}
\setlength\tabcolsep{5pt}
\begin{tabular}{l|ccc|ccc|ccc}
&  \multicolumn{3}{c}{\textsc{FiNER}} &  \multicolumn{3}{c}{ \textsc{G\"ung\"or-NN}}  &  \multicolumn{3}{c}{ \textsc{Sohrab-NN}}  \\ 
class & pre & rec & F1 & pre & rec & F1 & pre & rec & F1 \\
\hline
\noalign{\smallskip}
DATE & 97.30 & 96.52 & 96.91 &	80.49 & 52.04 & 63.21 &	98.88 & 61.37 & 75.74 \\
EVENT & 70.87 & 57.69 & 63.60 &	13.64 & 1.92 & 3.37 &	0.00 & 0.00 & 0.00 \\
LOC & 83.88 & 77.71 & 80.67 &	64.57 & 77.01 & 70.24 &	66.81 & 58.90 & 62.61 \\
ORG & 79.30 & 51.28 & 62.28 &	47.43 & 30.67 & 37.25 &	36.01 & 27.64 & 31.27 \\
PER & 85.32 & 77.79 & 81.38 &	82.15 & 68.79 & 74.88 &	72.03 & 46.52 & 56.53 \\
PRO & 73.80 & 49.32 & 59.12 &	30.28 & 40.79 & 34.76 &	28.30 & 31.35 & 29.75 \\
\hline
\noalign{\smallskip}
TOTAL & \bf 85.17 & \bf 72.47 & \bf 78.31 &	62.98 & 55.89 & 59.22 &	60.57 & 45.46 & 51.94 \\
\end{tabular}
\end{center}
\end{small}
\caption{Precision, recall, and F1-scores for top-level named entities in the Wikipedia test set. The highest overall precision, recall, and F1 values are bolded.}
\label{tab: results top-level wikipedia}
\end{table}

\subsection{Error Analysis}
\label{sec: error analysis}

Measured by F1-score over all name classes, the rule-based \textsc{FiNER} system outperformed the data-driven neural approaches \textsc{G\"ung\"or-NN} and \textsc{Sohrab-NN} on both the Digitoday (in-domain) and Wikipedia (out-of-domain) evaluation sets. The notably large performance gap between the approaches in the out-of-domain case can be explained by the fact that, as a rule-based system, \textsc{FiNER} incorporates a large amount of linguistic and world knowledge of its developer. Therefore, it is expected to generalize over varying domains rather well. Meanwhile, the trained neural approaches had ''observed'' non-Digitoday domains only implicitly via the pretrained word embeddings learned from all Finnish text available on the web (including Wikipedia) \citep{kanerva2014}, thus resulting in an understandably worse generalization capability. Nevertheless, on the in-domain Digitoday set, the  \textsc{G\"ung\"or-NN} provided a slightly better recall compared with \textsc{FiNER}. Therefore, as the in-domain recall is the ''weak spot'' of the \textsc{FiNER} system w.r.t. the neural approaches, we will next discuss it in more detail.

To this end we note that the recall of any singular class can be lowered due to four types of errors made by the NER system: 

\begin{itemize}
\item[1.] The system predicts the correct span of a gold standard entity but assigns an incorrect class.
\item[2.] The predicted entity has the correct class assignment but its span overlaps only partially with the gold standard entity.
\item[3.] The predicted entity overlaps only partially with the gold standard entity in addition to having an incorrect class assignment.
\item[4.] The system misses a gold standard entity completely.
\end{itemize}

While all these error types (referred to as ''recall errors'' from now on) contribute to the recall values identically, they are not necessarily equivalent from an application point of view. For example, it might be beneficial for a text summarizer to obtain even a partially recognized name: consider, e.g., ''Dreamworks'' versus ''Dreamworks Animation'' or ''Dreamworks Animation SKG''. In other words, some incorrect recognition results may in some situations be more acceptable than others.

The recall error counts yielded by \textsc{FiNER},  \textsc{G\"ung\"or-NN}, and  \textsc{Sohrab-NN} on the Digitoday test set are presented in Table \ref{tab: digitoday error analysis all entities} and on the Wikipedia test set in Table \ref{tab: wikipedia error analysis all entities}. We note that the distributions of the rule-based \textsc{FiNER} and the neural networks are different. Whereas most errors yielded by \textsc{FiNER} are of type 4 (66.6\% and 71.2\% for Digitoday and Wikipedia, respectively), the errors by  \textsc{G\"ung\"or-NN} and  \textsc{Sohrab-NN} are divided mostly between types 1 and 4. In other words, on the Digitoday test set, \textsc{FiNER}'s errors are mostly due to it completely missing entities, whereas the neural networks tend to predict the entity span correctly but then assign the entity an incorrect name class. This result might be interesting from the point of view of such applications which require a high recall of \textit{any} names ignoring the class labels, such as text anonymization \citep{kleinberg2017}. However, for this to make a difference in practice, the overall performance of the neural networks should naturally be improved to match that of the \textsc{FiNER} system.

\begin{table}[t!]
\begin{small}
\begin{center}
\begin{tabular}{l|ccccc}
model/type & 1 & 2 & 3 & 4 & TOTAL  \\
\hline
\noalign{\smallskip}
\textsc{FiNER} & 101 (11.5) & 131 (14.9) & 62 (7.1) &  586 (66.6) & 880 (100) \\
\textsc{G\"ung\"or-NN}&	271 (33.5) & 132 (16.3) & 68 (8.4) & 337 (41.7) & 808 (100) \\
 \textsc{Sohrab-NN}  &	426 (48.0) & 13 (14.6) & 9 (1.0) & 440 (49.6) & 888 (100) \\
\end{tabular}
\end{center}
\end{small}
\caption{The distributions of recall error types on the Digitoday test set. The absolute and relative frequencies are outside and inside the parentheses, respectively. See text for explanations of the Types 1-4.}
\label{tab: digitoday error analysis all entities}
\end{table}

\begin{table}[t!]
\begin{small}
\begin{center}
\begin{tabular}{l|ccccc}
model/type & 1 & 2 & 3 & 4 & TOTAL  \\
\hline
\noalign{\smallskip}
\textsc{FiNER} & 141 (8.0) & 201 (11.4) & 166 (9.4) & 1254 (71.2) & 1762 (100) \\
\textsc{G\"ung\"or-NN}& 795 (28.5) & 458 (16.4) & 314 (11.3) & 1221 (43.8) & 2788 (100) \\
 \textsc{Sohrab-NN}  & 1384 (42.6) & 34 (1.1) & 95 (2.9) & 1738 (53.5) & 3251 (100) \\
\end{tabular}
\end{center}
\end{small}
\caption{The distributions of recall error types on the Wikipedia test set. The absolute and relative frequencies are outside and inside the parentheses, respectively. See text for explanations of the Types 1-4.}
\label{tab: wikipedia error analysis all entities}
\end{table}

\section{Discussion}
\label{sec: discussion}

In this section, we discuss the choice of named entity classes and the NER systems employed in the experimental section of this paper in Sections \ref{sec: the named entity classes} and \ref{sec: previous work}, respectively.

\subsection{The Named Entity Classes}
\label{sec: the named entity classes}

The set of named entity classes included in an annotation scheme in general depends largely on the text domain. As for newswire text, it has most often been a priority to tag person, location, and organization names \citep{grishman1996,tjong2003,gustafson2006,benikova2014}. Therefore, considering the (technology) news domain of the Digitoday text, these were a natural inclusion also in our annotation scheme. 

While some previous work has assigned products similarly to their own separate category \citep{gustafson2006}, it appears to have been more common to treat them as a miscellaneous/other group of entities \citep{tjong2003,benikova2014}. This is most probably due to the scarcity of occurrences in the chosen text domains. Similarly, in \citep{tjong2003,benikova2014}, events were assigned to the miscellaneous/other category. In our work, we assigned both products and events to their separate entity classes. As for products, the separate class is easily justifiable since they occurred very frequently in both Digitoday and Wikipedia data (see Tables \ref{tab: counts of classes}, \ref{tab: digitoday test class counts}, and \ref{tab: wikipedia test class counts}). In contrast, events were clearly the least frequent class. In future experimental work employing the data sets described here one could consider discarding the event class altogether since there are not enough data to neither train nor evaluate models on that class. Nevertheless, from a human annotator's point of view, the event class as well as the other classes specified by our annotation guideline appear to be quite salient as demonstrated by the high inter-annotator agreement documented in Section \ref{sec: iaa}.

Dates, or more generally temporal expressions, have been incorporated in previous work to varying extents. For example, in the English and German CoNLL-2003 \citep{tjong2003} and the German NoSta-D \citep{benikova2014} data, temporal expressions are not considered.  Meanwhile, the MUC-6 and MUC-7 data sets take into account a wide range of temporal expressions, including absolute expressions such as ''3.10.2016'' and relative ones such as ''next week''. In addition, it should be noted, that recognition of time expressions and temporal relations has been a research topic of its own interest exemplified by the TempEval competition \citep{verhagen2007,verhagen2010,verhagen2013}.

Finally, during the first annotation stage of the corpus, we considered marking titles of people if they appeared in the immediate pre-context of proper names. For example, in the following, titles and their combinations would have been marked: \textbf{presidentti} Sauli Niinist\"o (\textbf{president} Sauli Niinist\"o), \textbf{perustaja ja varapuheen\-johtaja} Ville Oksanen (\textbf{founder and vice president} Ville Oksanen). However, we discarded this class during the second annotation stage since person titles are not a commonly included category in named entity annotation. They can be, however, found for example in related work on semantic webs \citep{ehrmann2017}.

\subsection{Previous Work}
\label{sec: previous work}

As described in Sections \ref{sec: corpus} and \ref{sec: finer}, the \textsc{FiNER} system and the Digitoday corpus have been under development at the University of Helsinki since 2013 and 2016, respectively. During this time, they have been publicly available and employed in a body of publications. In what follows, we summarize this previous work. 

The \textsc{FiNER} system was studied extensively in the 19th century Finnish newspaper domain \citep{kettunen2017a, kettunen2017b, kettunen2017c} with a particular focus on locations, person names, organization, and dates/time. The yielded results, however, were extremely poor. As pointed out by \citet{kettunen2017a}, the low results (F1 scores under 50 for all classes) were mainly due to two reasons: low optical character recognition accuracy of the historical material (only 70--75\% of words in data were correctly recognized) and the fact that \textsc{FiNER} was developed for modern Finnish.

In \citep{kettunen2017c}, \textsc{FiNER} was employed on the Digitoday corpus in addition to the historical domain. As for the Digitoday data, \citet{kettunen2017c} considered only three named entity classes: person names, locations, and organizations (coined corporations by them). Their reported results were much lower (F1 scores of 55.39,74.58, and 62.56 for person names, locations, and organizations, respectively) compared to the ones reported here. However, it is not possible to provide a meaningful comparison between these experiments since, at the time when their work took place, both the \textsc{FiNER} system and the Digitoday corpus were very much still under development.

Finally, \citet{gungor2018a} employed the Digitoday corpus as a part of their study on the effect of morphology on named entity recognition. Again, their experiments were run on a version under development. Therefore, their reported F1 scores are not comparable to the results presented here.

\subsection{On the Performance of the Data-Driven Systems}
\label{sec: the performance of data-driven systems}

As discussed in Section \ref{sec: finer}, the development of \textsc{FiNER} has resulted in a extensive gazetteers of well-known named entities, many of which have made it into the morphological analyser \textsc{OMorFi}. On the other hand, \textsc{OMorFi} is an integral part of the \textsc{FinnPos} morphological tagging and lemmatizing toolkit employed to preprocess the corpus for the data-driven \textsc{G\"ung\"or-NN} system employed in the experiments. Therefore, \textsc{G\"ung\"or-NN} does at least partially receive the \textsc{OMorFi} gazetteer information via the FinnPos tags (the recognized proper nouns receive the tag 'PROPN'). However, this apparently was not sufficient to enable the system to generalize from the Digitoday to the Wikipedia data set despite the system having access to world knowledge through pretrained word embeddings.

The poor results of the data-driven systems on the out-of-domain test set presented in Section \ref{sec: results} imply that the Digitoday corpus should be extended if the intention is to learn a general NER system for Finnish using data-driven methods despite the Digitoday corpus being comparable in size to other gold-standard NER datasets. We encourage creating more training data using a semi-automatic approach by first applying \textsc{FiNER} to a text collection (e.g. Wikipedia) and subsequently correcting the result manually. Given the evidently high accuracy of the rule-based \textsc{FiNER} system, one would expect this approach to be extremely fast compared with manually annotating raw text from scratch. The semi-automatically annotated data should then enable the data-driven systems to catch up with the accuracy of the rule-based \textsc{FiNER} tool. It is future work to determine if fully automated annotation of new material with \textsc{FiNER} without manual correction could in fact already help a data-driven system to adapt to a new domain.

\section{Conclusions}
\label{sec: conclusions}

We described a corpus consisting of Finnish technology related news articles with manually prepared named entity annotation. The corpus consists of 953 articles (193,742 word tokens) and is manually annotated using the nested annotation approach with a tag set comprising six NE tags (person, location, organization, product, event, date). The corpus is available for research purposes and can be readily used for development of NER systems for Finnish.

We described experimental results on three NER systems trained/developed on the corpus: a rule-based \textsc{FiNER} toolkit and two recently published neural network architectures. The systems were evaluated on two external data sets consisting of Digitoday and Wikipedia articles corresponding to in-domain and out-of-domain evaluation sets, respectively. The best performance on both Digitoday and Wikipedia evaluation sets was yielded by the rule-based \textsc{FiNER} toolkit with total F1-scores of 85.20 and 79.91, respectively.

\section{Acknowledgements}
\label{sec: acknowledgements}

We would like to express our sincere gratitude to Onur G\"ung\"or and Mohammad Golam Sohrab for their invaluable help with running the experiments. This work was funded by Academy of Finland (award numbers 292260 and 293239) and FIN-CLARIN. The third author has received funding from the European Research Council (ERC) under the European Union's Horizon 2020 research and innovation programme (grant agreement No 771113).  This is a post-peer-review, pre-copyedit version of an article published in Language Resources and Evaluation. The final authenticated version is available online at: http://dx.doi.org/10.1007/s10579-019-09471-7.

\newpage
\bibliographystyle{plainnat}

\end{document}